\documentclass[runningheads]{llncs}
\usepackage{comment}
\usepackage{amsmath,amssymb} 
\usepackage{color}

\usepackage{multirow}
\usepackage[ruled,vlined]{algorithm2e}
\usepackage{algorithmic}
\usepackage{algorithmic}
\usepackage{bm}
\usepackage{amsfonts,threeparttable,caption,tabularx,subfigure,graphicx,epstopdf,booktabs,longtable,supertabular}
\usepackage[pagebackref=true,breaklinks=true,letterpaper=true,colorlinks,bookmarks=false]{hyperref}

\begin{document}
\pagestyle{headings}
\mainmatter
\def\ECCVSubNumber{977}  

\title{SPL-MLL: Selecting Predictable Landmarks for Multi-Label Learning} 


%
\author{Junbing Li\inst{1} \and
	Changqing Zhang\inst{1,}\thanks{Corresponding author: Changqing Zhang.} \and
	Pengfei Zhu\inst{1} \and \\
	Baoyuan Wu\inst{2} \and
	Lei Chen\inst{3} \and
	Qinghua Hu\inst{1}	
}
%

\authorrunning{J. Li et al.}
%
\institute{Tianjin Key Lab of Machine Learning, College of Intelligence and Computing, Tianjin University, China  \and
The Chinese University of Hong Kong, Shenzhen, Tencent AI lab, China \and 
Nanjing University of Posts and Telecommunications, China\\
\email{\{lijunbing,zhangchangqing,zhupengfei,huqinghua\}@tju.edu.cn\\ wubaoyuan1987@gmail.com, chenlei@njupt.edu.cn}
}

\maketitle

\begin{abstract}
 Although significant progress achieved, multi-label classification is still challenging due to the complexity of correlations among different labels. Furthermore, modeling the relationships between input and some (dull) classes further increases the difficulty
of accurately predicting all possible labels. In this work, we propose to select a small subset of labels as landmarks which are easy to predict according to input (predictable) and can well recover the other possible labels (representative). Different from existing methods which separate the landmark selection and landmark prediction in the 2-step manner, the proposed algorithm, termed Selecting Predictable Landmarks for Multi-Label Learning (SPL-MLL), jointly conducts landmark selection, landmark prediction, and label recovery in a unified framework, to ensure both the representativeness and predictableness for selected landmarks. 
We employ the Alternating Direction Method (ADM) to solve our problem. Empirical studies on real-world datasets show that our method achieves superior classification performance over other state-of-the-art methods.

\end{abstract}

\section{Introduction}
Multi-label classification jointly assigns one sample with multiple tags reflecting its semantic content, which has been widely used in many real-world applications. In document classification, there are multiple topics for one document; in computer vision, one image may contain multiple types of object; in emotion analysis, there may be combined types of emotions, \emph{e.g.}, relaxing and quiet. Though plenty of multi-label classification methods \cite{tsoumakas2007multi,read2008multi,zhang2014review,read2011classifier,hou2016multi,liu2017deep,wu2017diverse,wu2018multi} have been proposed, multi-label classification is still a recognized challenging task due to the complexity of label correlations, and the difficulty of  predicting all labels.

\begin{figure}[t]
	\centering
	\includegraphics[width=3.6in, height=1.7in]{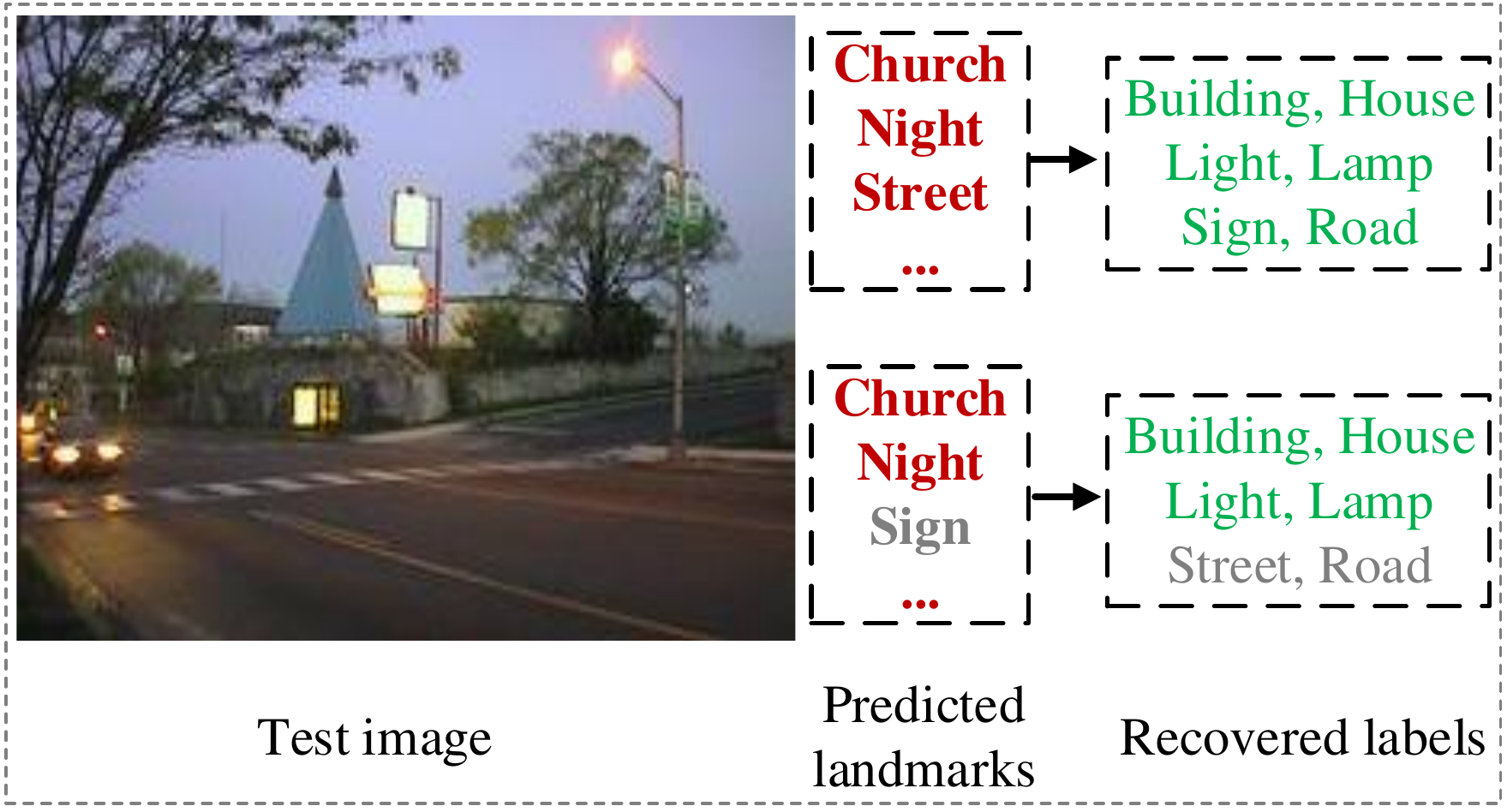}
	\caption{Why landmarks should be predictable? With a test image, although the other related labels (\emph{e.g.}, ``street", ``road") can be inferred from ``sign", the landmark label ``sign" itself is difficult to accurately predict (bottom row). While based on the image, the label ``street" is more predictable, and accordingly, more related labels are correctly inferred (top row). }
	\label{fig:coverFig1}
\end{figure}
In real-world applications, labels are usually correlated, and simultaneously predicting all possible labels is usually rather difficult. Accordingly, there are techniques aiming to reduce the label space. The representative strategy is landmark based multi-label classification. Landmark based methods first select a small subset of representative labels as landmarks, where the landmark labels are able to establish interdependence with other labels. The algorithm \cite{Balasubramanian2012The} employ group-sparse learning to select a few labels as landmarks which can reconstruct the other labels. However, this method divides the landmark selection and landmark prediction as two separate processes. The method \cite{bi2013efficient} performs label selection based on randomized sampling, where the sampling probability of each class label reflects its importance among all the labels. Another representative strategy is label embedding \cite{hsu2009multi,tai2012multilabel,chen2012feature,zhang2012maximum,zhou2012compressed}, which transforms label vectors into low-dimensional embeddings, where the correlations among labels can be implicitly encoded.

\begin{figure*}[]
	\centering
	\includegraphics[width=4.4in, height = 2.0in]{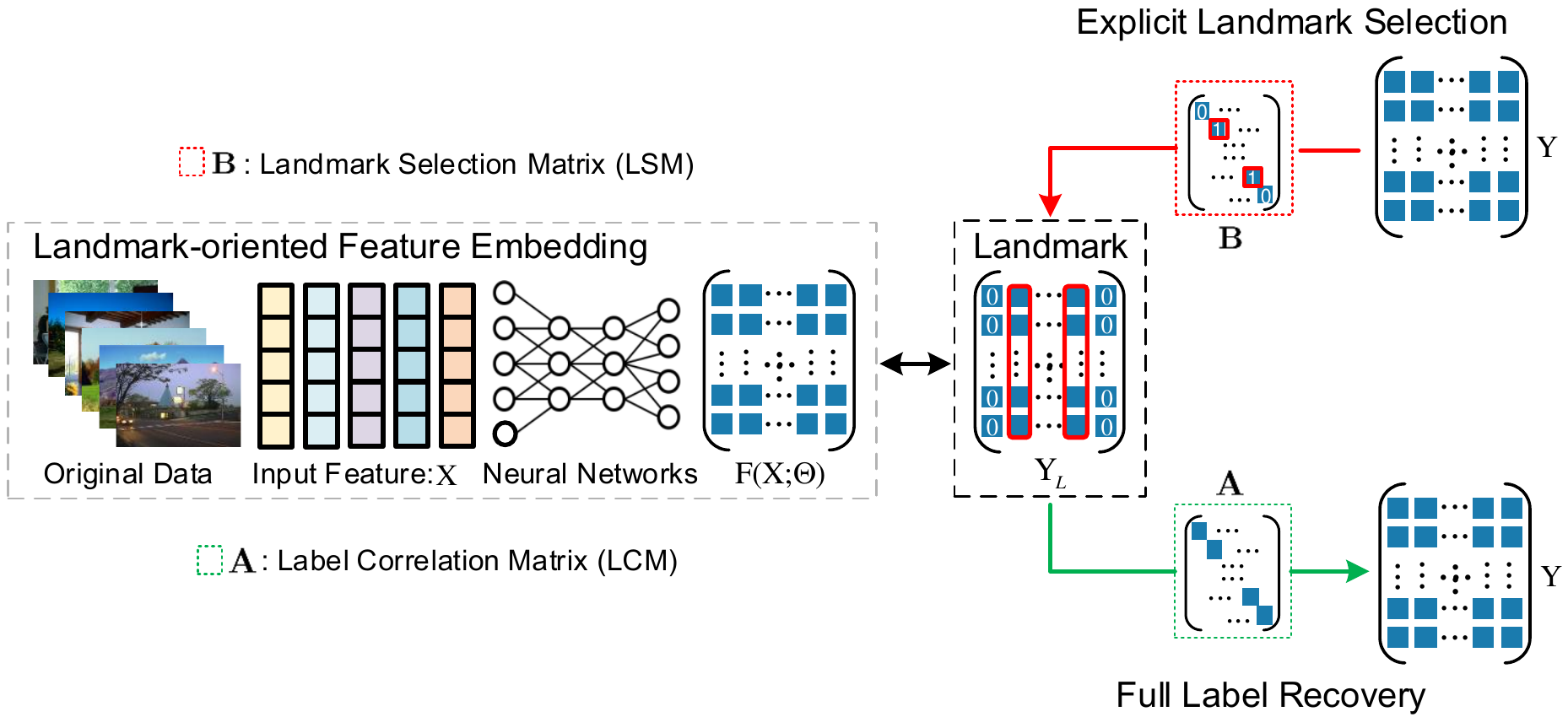}
	\caption{Overview of SPL-MLL. The key component of our model is the landmark selection strategy, which induces the explicit landmark label matrix $\mathbf{Y}_L$. The matrix $\mathbf{B}$, termed as  landmark selection matrix, is used to construct the landmark label matrix explicitly, while the matrix $\mathbf{A}$ is used to reconstruct all possible labels from landmarks. Benefitting from the explicit landmark label matrix $\mathbf{Y}_L$, the input is also able to be taken into account to ensure the predictable property for landmarks.}
	\label{fig:framework2}
\end{figure*}

Although several methods have been proposed to reduce the dimensionality of label space, there are several limitations left behind these methods. First, existing landmark-based methods usually separate the landmark label selection and landmark prediction into two independent steps. Therefore, even the other labels could be easily recovered from the selected landmarks, but the landmarks themselves may be difficult to be accurately predicted with input (as shown in Fig.~\ref{fig:coverFig1}). Second, the label-embedding-based methods usually project the original label space into a low-dimensional embedding space, where embedded vectors can be used to recover the full-label set. Although the embedded vectors may be easy to predict, the embedding way (\emph{i.e.}, dimensionality reduction) may cause information loss, and the label correlations are implicitly encoded thus lack interpretability. Considering the above issues, we jointly conduct landmarks selection, landmarks prediction and full-label recovery in a unified framework, and accordingly, propose a novel multi-label learning method, termed \emph{Selecting Predictable Landmarks for Multi-Label Learning} (\textbf{\emph{SPL-MLL}}). The overview of SPL-MLL is shown in Fig.~\ref{fig:framework2}.
The main advantages of the proposed algorithm include: (1) compared with existing landmark-based multi-label learning methods, SPL-MLL can select the landmarks which are both representative and predictable due to the unified objective; (2) compared with the embedding methods, SPL-MLL is more interpretable due to explicitly exploring the correlations with landmarks.

The contributions of this work are summarized as:
\begin{itemize}
	\item{We propose a novel landmark-based multi-label learning algorithm for complex correlations among labels. The landmarks bridge the intrinsic correlations among different labels, while also reduce the complexity of correlations and possible label noise.}
	\item{To the best of our knowledge, SPL-MLL is the first algorithm which simultaneously conducts landmark selection, landmark prediction, and full-label recovery in a unified objective, thus taking both representativeness and predictability for landmarks into account. This is quite different from the 2-step manner separating landmark selection and prediction.}
	\item{Extensive experiments on benchmark datasets are conducted, validating the effectiveness of the proposed method over state-of-the-arts.}
\end{itemize}

\section{Related Work}
Generally, existing multi-label methods can be roughly categorized into three lines based on the order of label correlations \cite{zhang2014review}. The first-order strategy \cite{boutell2004learning,zhang2007ml} tackles multi-label learning problem in the label-by-label manner, which ignores the co-existence of other labels. The second-order strategy \cite{elisseeff2002kernel,furnkranz2008multilabel,ghamrawi2005collective} conducts multi-label learning problem by introducing the pairwise relations between different labels. For high-order strategy \cite{ji2010shared,read2011classifier,tsoumakas2011random,wu2018tagging}, multi-label learning problem is solved by establishing more complicated label relationships, which makes these approaches tend to be quite computationally expensive.

In order to reduce label space, there are approaches based on label embedding, which searches a low-dimensional subspace so that correlations among labels can be implicitly expressed \cite{hsu2009multi,tai2012multilabel,chen2012feature,ren2019label,jia2019facial,zhang2012maximum,zhou2012compressed,yu2014large,lin2014multi,li2015multi}.
Based on the low-dimensional latent label space, one can effectively reduce computation cost while performing multi-label prediction. The representative embedding based methods include: label embedding via random projections \cite{hsu2009multi}, principal label space transformation (PLST) \cite{tai2012multilabel} and its conditional version (CPLST) \cite{chen2012feature}. Beyond considering linear embedding functions, there are several approaches employing standard kernel functions (\emph{e.g.}, low-degree polynomial kernels) for nonlinear label embedding. The work in \cite{yeh2017learning} proposes a novel DNN architecture of Canonical-Correlated Autoencoder (C2AE), which is a DNN-based label embedding framework for multi-label classification, which is able to perform feature-aware label embedding and label-correlation aware prediction.

To explore label correlations, there are several landmark based multi-label classification models aiming to reduce the label space \cite{Balasubramanian2012The,bi2013efficient,zhou2012compressed,boutsidis2009improved}. They usually first select a small subset of labels as landmarks, which are supposed to be representative and able to establish interdependency with other labels. The work in \cite{Balasubramanian2012The} models landmark selection with group-sparsity technique.
Following the assumption in \cite{Balasubramanian2012The}, the method in \cite{bi2013efficient} alleviates this problem of computation cost by proposing an valid label selection method based on randomized sampling, and utilizes the leverage score in the best rank-$\emph{k}$ subspace of the label matrix to obtain the sampling probability of each label.
It is noteworthy that these methods separate the landmark selection and landmark prediction in a 2-step manner, which can not simultaneously guarantee the representativeness and predictability of landmarks.

\vspace{-0.5cm}
\section{Our Algorithm: Selecting Predictable Landmarks for Multi-Label Learning}
For clarification, we first provide the definitions for symbols and variables used through out this paper. Let $\mathcal{X} = \mathbb{R}^{D} $ and $\mathcal{Y} = \{0,1\}^{C} $ denote the feature space and label space, where $D$ and $C$  are the dimensionality of feature space and label space, respectively. Given training data with the form of instance-label pairs $\{{{\mathbf{x}}_i},{{\mathbf{y}}_i}\} _{i = 1}^N $, accordingly, the feature matrix can be represented as $\mathbf{X} \in \mathbb{R}^{N \times D}$, and the label matrix is represented as $ \mathbf{Y} \in \mathbb{R}^{N \times C}$. The goal of multi-label learning is to learn a model $f:\mathcal{X} \to \mathcal{Y}$, to predict possible labels accurately for new coming instances. Motivated by the landmark strategy, we propose a novel algorithm for multi-label learning, termed SPL-MLL, \emph{i.e.}, \emph{Selecting Predictable Landmarks for Multi-Label Learning}. SPL-MLL consists of two key components, \emph{i.e.}, \emph{Explicit Landmark Selection} and \emph{Predictable Landmark Classification}. 

\subsection{Explicit Landmark Selection}
Different from the 2-step manner \cite{Balasubramanian2012The} which only focuses on selecting landmarks that are most representative, our goal is to select landmarks which are both representative and predictable. There are two designed matrixes which are the keys to realize this goal. The first matrix is the \textbf{\emph{label correlation matrix (LCM)}} $\mathbf{A}$ used for recovering other labels with landmarks. In self-representation manner, the matrix $\mathbf{A} \in \mathbb{R}^{C \times C}$ is obtained which captures the correlation among labels and explores the interdependency between landmark labels and the others. In the work \cite{Balasubramanian2012The}, the landmarks are selected implicitly. Specifically, the underlying assumption is ${\mathbf{Y}}= \mathbf{YA}$, where $\mathbf{A}$ is constrained by minimizing $||\mathbf{A}||_{2,1}$ to enforce the reconstruction of $\mathbf{Y}$ mainly base on a few labels, \emph{i.e.}, landmarks. In our model, although the linear self-representation manner is also introduced in our model, we try to obtain the landmark label matrix explicitly in the objective function.

The second critical matrix is the \textbf{\emph{landmark selection matrix (LSM)}} $\mathbf{B}$. Note that, in the work \cite{Balasubramanian2012The}, there is no explicit landmark label matrix constructed and the selection result is implicitly encoded in $\mathbf{A}$ due to its sparsity in row. Both $\mathbf{YA}$ and $\mathbf{Y}$ ($\mathbf{YA}\approx \mathbf{Y}$) in \cite{Balasubramanian2012The} are full-label matrix. Different from \cite{Balasubramanian2012The}, since we aim to jointly conduct landmark selection and learn a model to predict these selected landmarks instead of all labels, we need to explicitly derive a label matrix encoding the landmarks. To this end, we introduce the matrix $\mathbf{B} \in \mathbb{R}^{C \times C}$ which is a diagonal matrix, and each diagonal element is either 0 or 1, \emph{i.e.}, ${B}_{ii} \in \{0,1\}$. Then, we can obtain the explicit landmark label matrix ${\mathbf{Y}}_L$ with ${\mathbf{Y}}_L = \mathbf{YB}$. In this way, the columns corresponding to the landmarks in $\mathbf{Y}_L$ unchanged while the elements of other columns (corresponding to non-landmark labels) will be 0. It is noteworthy that $\mathbf{B}$ is learned in our model instead of being fixed in advance. Accordingly, the explicit landmark selection objective to minimize is induced as:
\begin{equation}
	\begin{aligned}
		\mathbf{\Gamma}(\mathbf{B},\mathbf{A})&=\|\mathbf{Y}-\mathbf{Y}_{L}\mathbf{A}\|_{F}^2+\mathbf{\Omega}(\mathbf{B})\\
		&=\|\mathbf{Y}-\mathbf{Y}\mathbf{B}\mathbf{A}\|_{F}^2+\mathbf{\Omega}(\mathbf{B}),\\
		&s.t.~B_{ij} = 0, ~i \neq j; ~B_{ij} \in \{0,1\}, ~i = j.
	\end{aligned}
	\label{loss}
\end{equation}
Since it is difficult to strictly ensure the diagonal property for $\mathbf{B}$, a soft constraint $\mathbf{\Omega}(\mathbf{B})$ is introduced as follows:
\begin{equation}
	\begin{aligned}
		\mathbf{\Omega}(\mathbf{B})&= \lambda_1\|\mathbf{B}-\mathbf{I}\|_{F}^2+\lambda_2\|\mathbf{B}\|_{2,1},\\
	\end{aligned}
	\label{loss}
\end{equation}
where the structure sparsity $ ||\mathbf{B}||_{2,1}=\sum_{i=1}^{C}\sqrt{\sum_{j=1}^{C}B_{ij}^2}$ is used to select a few landmarks, and the approximation to the identity matrix $\mathbf{I}$ ensures the labels corresponding to landmarks unchanged. The regularization parameter $\lambda_1$ and $\lambda_2$  control the degree of diagonal and sparsity property for $\mathbf{B}$, respectively. 
Notice that the label correlation matrix $\mathbf{A}$ is learned automatically without constraint, the underlying assumption for the correlation is sparse (similar to the existing work \cite{Balasubramanian2012The}) which is jointly ensured by the sparse landmark selection matrix $\mathbf{B}$.
Then, we can obtain the explicit landmark label matrix $\mathbf{Y}_L = \mathbf{YB}$, and train a prediction model exactly for the landmarks.

\subsection{Predictable Landmark Classification}
Now, we firstly consider learning the classification model for accurately predicting landmarks instead of all possible labels. Beyond label correlation, modeling $\mathcal{X} \to \mathcal{Y}$ is also critical in multi-label classification. However, the traditional landmark-based multi-label classification algorithms usually separate landmark selection and landmark prediction, which may result in unpromising classification accuracy because the selected landmarks may be representative but difficult to be predicted (see Fig.~\ref{fig:coverFig1}). Recall that the goal of our model is to recover full labels with landmark labels, so our classification model only focuses on predicting landmarks $\mathbf{Y}_L$ based on $\mathbf{X}$ instead of full labels $\mathbf{Y}$. Accordingly, our predictable landmark classification objective to minimize is as follows:
\begin{equation}
	\begin{aligned}
		\mathbf{\Phi}(\mathbf{B},\mathbf{\Theta})&=\|\bm{f}(\mathbf{X};\mathbf{\Theta})\mathbf{B}-\mathbf{Y}_{L}\|_{F}^2\\
		&=\|(\bm{f}(\mathbf{X};\mathbf{\Theta})-\mathbf{Y})\mathbf{B}\|_{F}^2,
	\end{aligned}
	\label{loss}
\end{equation}
where $\bm{f}(\cdot;\mathbf{\Theta})$ is the neural networks (parameterized by $\mathbf{\Theta}$) used for feature embedding and conducting classification for landmarks, which is implemented by fully connected neural networks.

\subsection{Objective Function}
Based on above considerations, a novel landmark-based multi-label classification algorithm, \emph{i.e.}, Selecting Predictable Landmarks for Multi-Label Learning (SPL-MLL), is induced, which jointly learns landmark selection matrix, label correlation matrix, and landmark-oriented feature embedding in a unified framework. Specifically, the objective function of SPL-MLL for us to minimize is as follows:
\begin{equation}
	\begin{aligned}
		\mathcal{L}(\mathbf{B},&\mathbf{\Theta},\mathbf{A})=\mathbf{\Phi}(\mathbf{B},\mathbf{\Theta})+\mathbf{\Gamma}(\mathbf{B},\mathbf{A})\\
		&=\|(\bm{f}(\mathbf{X};\mathbf{\Theta})-\mathbf{Y})\mathbf{B}\|_{F}^2\\
		&+\|\mathbf{Y}-\mathbf{Y}\mathbf{B}\mathbf{A}\|_{F}^2+\lambda_1\|\mathbf{B}-\mathbf{I}\|_{F}^2+\lambda_2\|\mathbf{B}\|_{2,1}.\\
	\end{aligned}
	\label{loss-function}
\end{equation}
It is noteworthy that the critical role of matrix $\mathbf{B}$, which bridges the landmark selection and landmark classification model. With this strategy, the proposed model jointly selects predictable landmark labels, captures the correlations among labels, and discovers the nonlinear correlations between features and landmarks, accordingly, promotes the performance of multi-label prediction.

\subsection{Optimization}
Since the objective function of our SPL-MLL is not jointly convex for all the variables, we optimize our objective function by employing Alternating Direction Minimization(ADM) \cite{lin2011linearized} strategy. To optimize the objective function in Eq.~(\ref{loss-function}), we should solve three subproblems with respect to $\mathbf{\Theta}$, $\mathbf{B}$ and $\mathbf{A}$, respectively. The optimization is cycled over updating different blocks of variables. 
We apply the technique of stochastic gradient descent for updating $\mathbf{\Theta}$, $\mathbf{B}$ and $\mathbf{A}$. The details of optimization are demonstrated as follows:\\
$\bullet$ Update networks. The back-propagation algorithm is employed to update the network parameters.\\
$\bullet$ Update $\mathbf{B}$. The gradient of $\mathcal{L}$ with respect to $\mathbf{B}$ can be derived as:
\begin{equation}
	\begin{aligned}
		\frac{\partial{\mathcal{L}}}{\partial{\mathbf{B}}}&=2(\bm{f}(\mathbf{X};\mathbf{\Theta})-\mathbf{Y})^T(\bm{f}(\mathbf{X};\mathbf{\Theta})-\mathbf{Y})\mathbf{B}\\
		&-2\mathbf{Y}^T(\mathbf{Y}-\mathbf{Y}\mathbf{B}\mathbf{A})\mathbf{A}^T+2\lambda_1(\mathbf{B}-\mathbf{I})+2\lambda_2\mathbf{D}\mathbf{B},
	\end{aligned}
	\label{gradient-B}
\end{equation}
where $\mathbf{D}$ is a diagonal matrix with $D_{ii}=\frac{1}{{2||{\mathbf{B}_i}||}}$. Accordingly, gradient descent is employed based on Eq.~(\ref{gradient-B}). \\
$\bullet$ Update $\mathbf{A}$. The gradient of $\mathcal{L}$ with respect to $\mathbf{A}$ can be derived as:
\begin{equation}
	\begin{aligned}
		\frac{\partial{\mathcal{L}}}{\partial{\mathbf{A}}}&=-2\mathbf{B}^T\mathbf{Y}^T(\mathbf{Y}-\mathbf{Y}\mathbf{B}\mathbf{A}).\\
	\end{aligned}
	\label{gradient-A}
\end{equation}
then $\mathbf{A}$ is updated by applying gradient descent based on Eq.~(\ref{gradient-A}).
The optimization procedure of SPL-MLL is summarized as Algorithm 1.

Once the model of SPL-MLL is obtained, it can be easily applied for predicting the labels of test samples. Specifically, given a test input $\mathbf{x}$, it  will be first transformed into $\bm{f}(\mathbf{x};\mathbf{\Theta})$, followed by utilizing the learned mappings $\mathbf{B}$ and $\mathbf{A}$ to predict its all possible labels with $\mathbf{y} = \bm{f}(\mathbf{x};\mathbf{\Theta}) \mathbf{B}\mathbf{A}$.

\begin{algorithm}[t]
	\SetAlgoLined
	\caption{Algorithm of SPL-MLL}
	\KwIn{Feature matrix $ \mathbf{X} \in \mathbb{R}^{N \times D}$, label matrix  $ \mathbf{Y} \in \mathbb{R}^{N \times C}$, parameters $\lambda_1 , \lambda_2$.}
	\textbf{Initialize:} $ \mathbf{B}=\mathbf{I}$,   initialize randomly $ \mathbf{A} $.\\
	\While{not converged}{
		Update the parameters $\mathbf{\Theta}$ of $\bm{f}(\cdot;\mathbf{\Theta})$;\\
		Update $\mathbf{B}$ by Eq.(\ref{gradient-B});\\
		Update $\mathbf{A}$ by Eq.(\ref{gradient-A});\\
	}
	\KwOut{$\bm{f}(\cdot;\mathbf{\Theta}),\mathbf{B},\mathbf{A} $.}
	\label{alg:alg1}
\end{algorithm}

\section{Experiments}
\subsection{Experiment Settings}
We conduct experiments on the following benchmark multi-label datasets: emotions \cite{trohidis2008multi}, yeast \cite{elisseeff2002kernel}, tmc2007 \cite{chartemultilabel}, scene \cite{boutell2004learning}, espgame \cite{von2004labeling} and pascal VOC 2007 \cite{everingham2010pascal}. Specifically, emotions and yeast are used for music and gene functional classification, respectively; tmc2007 is a large-scale text dataset, while scene, espgame and pascal voc 2007 belong to the domain of image.
The description of features for emotions, yeast, tmc2007 and scene could be referred in \cite{trohidis2008multi,elisseeff2002kernel,chartemultilabel,boutell2004learning}.
For espgame and pascal voc 2007, the local descriptor DenseSift \cite{lowe2004distinctive} is used. These datasets can be found in Mulan \footnote[1]{http://mulan.sourceforge.net/datasets-mlc.html\label{footnote1}} and LEAR websites \footnote[2]{http://lear.inrialpes.fr/people/guillaumin/data.php\label{footnote2}}. The detailed statistics information of each dataset is listed in Table \ref{tab:addlabel1}. We employ the standard partitions for training and testing sets \textsuperscript{\ref {footnote1},\ref {footnote2}}.

For the proposed SPL-MLL, we utilize neural networks for feature embedding and classification. The networks consists of 2 layers: for the first and second fully connected layer, 512 and 64 neurons are deployed, respectively. A leaky ReLU activation function is employed with the batch size being 64. In addition, we initialize the matrix $\mathbf{B}$ with $\mathbf{B}=\mathbf{I}$ which captures the most sparse correlation among labels and is beneficial to landmark selection. The regularization parameters, i.e., $\lambda_1$ and $\lambda_2$ are both fixed as 0.1 for all datasets and promising performance is obtained. In our experiments, we set the constraint $\mathbf{B}_{ij}=0,i \neq j$ in each iteration of optimization. This strictly guarantees the diagonal property and can provide clear interpretability for landmarks. The experimental results show that both convergence of our model and promising performance are achieved with this constraint.
\vspace{-0.5cm}
\begin{table}[htbp]
	\centering
	\caption{Statistics of datasets.}
	\setlength{\tabcolsep}{2.5mm}{	
		\begin{tabular}{cccccc}
			\hline
			dataset & \#instances & \#features & \#labels & cardinality & domain \\
			\hline
			emotions & 593   & 72    & 6     & 1.9   & music \\
			scene & 2407  & 294   & 6     & 1.1   & image \\
			yeast & 2417  & 103   & 14    & 4.2   & biology \\
			tmc2007 & 28596 & 500   & 22    & 2.2   & text \\
			espgame & 20770 & 1000  & 268   & 4.7   & image \\
			pascal VOC 2007 & 9963 & 1000  & 20   & 1.5   & image \\
			\hline
	\end{tabular}}%
	\label{tab:addlabel1}%
\end{table}%
Five diverse metrics are employed for performance evaluation. For \emph{Hamming loss} and \emph{Ranking loss}, smaller value indicates better classification quality, while larger value of \emph{Average precision}, \emph{Macro-F1} and \emph{Micro-F1} means better performance. These evaluation metrics evaluate the performance of multi-label predictor from various aspects, and details of these evaluation metrics can be found in \cite{zhang2014review}. 10-fold cross-validation is performed for each method, which randomly holds 1/10 of training data for validation during each fold. We repeat each experiment 10 times and report the averaged results with standard derivations.

\subsection{Experimental Results}

\subsubsection{Comparison with state-of-the-art multi-label classification methods}
\begin{table*}[!ht]\small
	\centering
	\renewcommand\arraystretch{0.75}
	\caption{ Comparing results (mean $\pm$ std.) of multi-label learning algorithms. $\downarrow$ ($\uparrow$) indicates the smaller (larger), the better. The values in red and blue indicate the best and the second best performances, respectively. $\bullet$ indicates that ours is better than the compared algorithms.}
	\resizebox{4.8in}{!}{
	\begin{tabular}{ccccccc}
		\toprule[1pt]
		Datasets & Methods & Ranking Loss $\downarrow$ & Hamming Loss $\downarrow$ & Average Precision $\uparrow$ & Micro-F1 $\uparrow$ & Macro-F1 $\uparrow$ \\
		\midrule[0.5pt]
		\multirow{11}[10]{*}{emotions} & BR \cite{tsoumakas2007multi}    & 0.309$\pm$0.021$\bullet$ & 0.265$\pm$0.015$\bullet$ & 0.687$\pm$0.017$\bullet$ & 0.592$\pm$0.025$\bullet$ & 0.590$\pm$0.016$\bullet$ \\
		& LP \cite{boutell2004learning}    & 0.345$\pm$0.022$\bullet$ & 0.277$\pm$0.010$\bullet$ & 0.661$\pm$0.018$\bullet$ & 0.533$\pm$0.016$\bullet$ & 0.504$\pm$0.019$\bullet$ \\
		& ML-kNN \cite{zhang2007ml} & 0.173$\pm$0.015$\bullet$ & 0.209$\pm$0.021$\bullet$ & 0.794$\pm$0.016$\bullet$ & 0.650$\pm$0.031$\bullet$ & 0.607$\pm$0.033$\bullet$ \\
		& EPS \cite{read2008multi}  & 0.183$\pm$0.014$\bullet$ & 0.208$\pm$0.010$\bullet$ & 0.780$\pm$0.017$\bullet$ & 0.664$\pm$0.012$\bullet$ & 0.655$\pm$0.018$\bullet$ \\
		& ECC \cite{read2011classifier}   & 0.198$\pm$0.021$\bullet$ & 0.228$\pm$0.022$\bullet$ & 0.766$\pm$0.014$\bullet$ & 0.617$\pm$0.013$\bullet$ & 0.597$\pm$0.019$\bullet$ \\
		& RAkEL \cite{tsoumakas2011random} & 0.217$\pm$0.026$\bullet$ & 0.219$\pm$0.013$\bullet$ & 0.766$\pm$0.031$\bullet$ & 0.634$\pm$0.023$\bullet$ & 0.618$\pm$0.036$\bullet$ \\
		& CLR \cite{furnkranz2008multilabel}   & 0.199$\pm$0.024$\bullet$ & 0.255$\pm$0.012$\bullet$ & 0.762$\pm$0.024$\bullet$ & 0.614$\pm$0.037$\bullet$ & 0.601$\pm$0.038$\bullet$ \\
		& MLML \cite{hou2016multi}  & 0.184$\pm$0.015$\bullet$ & 0.197$\pm$0.013$\bullet$ & 0.719$\pm$0.018$\bullet$ & 0.661$\pm$0.039$\bullet$  & 0.650$\pm$0.047$\bullet$ \\
		& MLFE \cite{zhang2018feature}    & 0.181$\pm$0.012$\bullet$ & 0.217$\pm$0.020$\bullet$ & 0.782$\pm$0.013$\bullet$ & 0.674$\pm$0.026$\bullet$ & 0.663$\pm$0.021$\bullet$ \\
		& HNOML \cite{zhang2019hybrid}   & 0.173$\pm$0.012$\bullet$ & 0.192$\pm$0.005$\bullet$ & 0.784$\pm$0.011$\bullet$ & 0.672$\pm$0.014$\bullet$ & 0.660$\pm$0.029$\bullet$ \\
		& Ours (linear)  & \textcolor{blue}{0.172$\pm$0.006}$\hspace*{1.6mm}$ & \textcolor{blue}{0.184$\pm$0.015}$\hspace*{1.6mm}$ & \textcolor{blue}{0.798$\pm$0.011}$\hspace*{1.6mm}$ & \textcolor{blue}{0.686$\pm$0.013}$\hspace*{1.6mm}$ & \textcolor{blue}{0.675$\pm$0.031}$\hspace*{1.6mm}$ \\						
		& Ours  & \textcolor{red}{0.170$\pm$0.004}$\hspace*{1.6mm}$ & \textcolor{red}{0.175$\pm$0.021}$\hspace*{1.6mm}$ & \textcolor{red}{0.815$\pm$0.014}$\hspace*{1.6mm}$ & \textcolor{red}{0.698$\pm$0.021}$\hspace*{1.6mm}$ & \textcolor{red}{0.687$\pm$0.024}$\hspace*{1.6mm}$ \\
		\hline
		\multirow{11}[10]{*}{yeast} & BR \cite{tsoumakas2007multi}    & 0.322$\pm$0.011$\bullet$ & 0.253$\pm$0.004$\bullet$ & 0.614$\pm$0.008$\bullet$ & 0.569$\pm$0.014$\bullet$ & 0.386$\pm$0.011$\bullet$ \\
		& LP \cite{boutell2004learning}    & 0.408$\pm$0.008$\bullet$ & 0.282$\pm$0.005$\bullet$ & 0.566$\pm$0.008$\bullet$ & 0.519$\pm$0.023$\bullet$ & 0.361$\pm$0.025$\bullet$ \\
		& ML-kNN \cite{zhang2007ml} & \textcolor{blue}{0.171$\pm$0.006}$\hspace*{1.6mm}$ & 0.218$\pm$0.004$\bullet$ & 0.757$\pm$0.011$\bullet$ & 0.636$\pm$0.012$\bullet$ & 0.357$\pm$0.021$\bullet$ \\
		& EPS \cite{read2008multi}   & 0.205$\pm$0.003$\bullet$ & 0.214$\pm$0.005$\bullet$ & 0.731$\pm$0.017$\bullet$ & 0.625$\pm$0.015$\bullet$ & 0.372$\pm$0.014$\bullet$ \\
		& ECC \cite{read2011classifier}   & 0.187$\pm$0.007$\bullet$ & 0.209$\pm$0.009$\bullet$ & 0.745$\pm$0.012$\bullet$ & 0.618$\pm$0.013$\bullet$ & 0.369$\pm$0.017$\bullet$ \\
		& RAkEL \cite{tsoumakas2011random} & 0.250$\pm$0.005$\bullet$ & 0.232$\pm$0.005$\bullet$ & 0.710$\pm$0.009$\bullet$ & 0.632$\pm$0.009$\bullet$ & 0.430$\pm$0.012$\bullet$ \\
		& CLR \cite{furnkranz2008multilabel}  & 0.187$\pm$0.005$\bullet$ & 0.222$\pm$0.005$\bullet$ & 0.745$\pm$0.008$\bullet$ & 0.628$\pm$0.012$\bullet$ & 0.400$\pm$0.018$\bullet$ \\
		& MLML \cite{hou2016multi}  & 0.178$\pm$0.002$\bullet$ & 0.224$\pm$0.005$\bullet$ & 0.757$\pm$0.009$\bullet$ & 0.641$\pm$0.014$\bullet$ & 0.443$\pm$0.025$\bullet$ \\
		& MLFE \cite{zhang2018feature}   & \textcolor{red}{0.169$\pm$0.021}$\hspace*{1.6mm}$ & 0.227$\pm$0.010$\bullet$ & 0.754$\pm$0.012$\bullet$ & 0.646$\pm$0.013$\bullet$ & 0.415$\pm$0.011$\bullet$ \\
		& HNOML \cite{zhang2019hybrid}    & 0.179$\pm$0.007$\bullet$ & 0.222$\pm$0.004$\bullet$ & 0.757$\pm$0.011$\bullet$ & 0.648$\pm$0.006$\bullet$ & 0.421$\pm$0.016$\bullet$ \\
		& Ours (linear)  & 0.172$\pm$0.003$\hspace*{1.6mm}$ & \textcolor{blue}{0.210$\pm$0.008}$\hspace*{1.6mm}$ & \textcolor{blue}{0.769$\pm$0.006}$\hspace*{1.6mm}$ & \textcolor{blue}{0.659$\pm$0.012}$\hspace*{1.6mm}$ & \textcolor{blue}{0.443$\pm$0.016}$\hspace*{1.6mm}$ \\					
		& Ours  & \textcolor{blue}{0.171$\pm$0.004}$\hspace*{1.6mm}$ & \textcolor{red}{0.201$\pm$0.006}$\hspace*{1.6mm}$ & \textcolor{red}{0.786$\pm$0.005}$\hspace*{1.6mm}$ & \textcolor{red}{0.667$\pm$0.011}$\hspace*{1.6mm}$ & \textcolor{red}{0.451$\pm$0.023}$\hspace*{1.6mm}$ \\
		\hline
		\multirow{11}[10]{*}{scene} & BR \cite{tsoumakas2007multi}    & 0.236$\pm$0.017$\bullet$ & 0.136$\pm$0.004$\bullet$ & 0.715$\pm$0.011$\bullet$ & 0.609$\pm$0.014$\bullet$ & 0.616$\pm$0.025$\bullet$ \\
		& LP \cite{boutell2004learning}    & 0.219$\pm$0.010$\bullet$ & 0.149$\pm$0.006$\bullet$ & 0.722$\pm$0.010$\bullet$ & 0.585$\pm$0.016$\bullet$ & 0.592$\pm$0.011$\bullet$ \\
		& ML-kNN \cite{zhang2007ml} & 0.093$\pm$0.009$\bullet$ & 0.095$\pm$0.008$\bullet$ & 0.851$\pm$0.016$\bullet$ & 0.718$\pm$0.015$\bullet$ & 0.719$\pm$0.024$\bullet$ \\
		& EPS \cite{read2008multi}   & 0.113$\pm$0.007$\bullet$ & 0.103$\pm$0.017$\bullet$ & 0.825$\pm$0.013$\bullet$ & 0.686$\pm$0.018$\bullet$ & 0.688$\pm$0.018$\bullet$ \\
		& ECC \cite{read2011classifier}   & 0.103$\pm$0.010$\bullet$ & 0.104$\pm$0.012$\bullet$ & 0.832$\pm$0.015$\bullet$ & 0.668$\pm$0.017$\bullet$ & 0.671$\pm$0.016$\bullet$ \\
		& RAkEL \cite{tsoumakas2011random} & 0.106$\pm$0.005$\bullet$ & 0.106$\pm$0.005$\bullet$ & 0.829$\pm$0.007$\bullet$  & 0.636$\pm$0.023$\bullet$ & 0.644$\pm$0.019$\bullet$ \\
		& CLR \cite{furnkranz2008multilabel}   & 0.106$\pm$0.003$\bullet$ & 0.138$\pm$0.003$\bullet$ & 0.817$\pm$0.006$\bullet$ & 0.612$\pm$0.026$\bullet$ & 0.620$\pm$0.025$\bullet$ \\
		& MLML \cite{hou2016multi}  & 0.079$\pm$0.004$\bullet$ & 0.098$\pm$0.013$\bullet$ & \textcolor{blue}{0.862$\pm$0.010}$\bullet$ & 0.728$\pm$0.029$\bullet$ & 0.729$\pm$0.029$\bullet$ \\
		& MLFE \cite{zhang2018feature}   & 0.079$\pm$0.002$\bullet$ & 0.094$\pm$0.003$\bullet$ & 0.858$\pm$0.013$\bullet$ & 0.732$\pm$0.021$\bullet$ & 0.734$\pm$0.019$\bullet$ \\
		& HNOML \cite{zhang2019hybrid}   & 0.103$\pm$0.005$\bullet$ & 0.110$\pm$0.003$\bullet$ & 0.832$\pm$0.108$\bullet$ & 0.733$\pm$0.011$\bullet$ & 0.736$\pm$0.013$\bullet$ \\
		& Ours (linear)  & \textcolor{blue}{0.073$\pm$0.003}$\hspace*{1.6mm}$ & \textcolor{blue}{0.083$\pm$0.006}$\hspace*{1.6mm}$ & 0.861$\pm$0.005$\hspace*{1.6mm}$ & \textcolor{blue}{0.738$\pm$0.012}$\hspace*{1.6mm}$ & \textcolor{blue}{0.742$\pm$0.021}$\hspace*{1.6mm}$ \\	
		& Ours  & \textcolor{red}{0.067$\pm$0.003}$\hspace*{1.6mm}$ & \textcolor{red}{0.074$\pm$0.004}$\hspace*{1.6mm}$ & \textcolor{red}{0.884$\pm$0.005}$\hspace*{1.6mm}$ & \textcolor{red}{0.746$\pm$0.016}$\hspace*{1.6mm}$ & \textcolor{red}{0.753$\pm$0.024}$\hspace*{1.6mm}$ \\
		\hline
		\multirow{11}[10]{*}{espgame} & BR \cite{tsoumakas2007multi}    & 0.266$\pm$0.003$\bullet$ & 0.019$\pm$0.002$\bullet$ & 0.221$\pm$0.001$\bullet$ & 0.205$\pm$0.004$\bullet$ & 0.116$\pm$0.001$\bullet$ \\
		& LP \cite{boutell2004learning}    & 0.496$\pm$0.003$\bullet$ & 0.031$\pm$0.001$\bullet$ & 0.055$\pm$0.004$\bullet$ & 0.109$\pm$0.003$\bullet$ & 0.060$\pm$0.002$\bullet$ \\
		& ML-kNN \cite{zhang2007ml} & 0.238$\pm$0.001$\bullet$ & \textcolor{blue}{0.017$\pm$0.002}$\hspace*{1.6mm}$ & 0.255$\pm$0.003$\bullet$ & 0.039$\pm$0.002$\bullet$ & 0.020$\pm$0.001$\bullet$ \\
		& EPS \cite{read2008multi}   & 0.380$\pm$0.001$\bullet$ & \textcolor{blue}{0.017$\pm$0.001}$\hspace*{1.6mm}$ & 0.200$\pm$0.003$\bullet$ & 0.083$\pm$0.002$\bullet$ & 0.065$\pm$0.001$\bullet$ \\
		& ECC \cite{read2011classifier}  & 0.230$\pm$0.001$\bullet$ & 0.020$\pm$0.002$\bullet$  & 0.282$\pm$0.001$\bullet$ & 0.245$\pm$0.004$\bullet$ & 0.123$\pm$0.001$\bullet$ \\
		& RAkEL \cite{tsoumakas2011random} & 0.343$\pm$0.001$\bullet$ & 0.019$\pm$0.001$\bullet$ & 0.211$\pm$0.003$\bullet$ & 0.150$\pm$0.003$\bullet$ & 0.059$\pm$0.001$\bullet$ \\
		& CLR \cite{furnkranz2008multilabel}   & \textcolor{red}{0.196$\pm$0.001}$\hspace*{1.6mm}$ & 0.019$\pm$0.001$\bullet$ & \textcolor{red}{0.305$\pm$0.003}$\hspace*{1.6mm}$ & 0.266$\pm$0.004$\bullet$ & 0.143$\pm$0.001$\bullet$ \\
		& MLML \cite{hou2016multi}  & 0.317$\pm$0.000$\bullet$ & 0.019$\pm$0.003$\bullet$ & 0.086$\pm$0.002$\bullet$ & 0.103$\pm$0.003$\bullet$ & 0.060$\pm$0.002$\bullet$ \\
		& MLFE \cite{zhang2018feature}   & 0.312$\pm$0.012$\bullet$ & 0.020$\pm$0.001$\bullet$ & 0.268$\pm$0.011$\bullet$ & 0.260$\pm$0.003$\bullet$ & 0.134$\pm$0.004$\bullet$ \\
		& HNOML \cite{zhang2019hybrid}   & 0.221$\pm$0.001$\bullet$ & 0.019$\pm$0.003$\bullet$ & 0.271$\pm$0.003$\bullet$ & 0.263$\pm$0.006$\bullet$ & 0.132$\pm$0.004$\bullet$ \\
		& Ours (linear)  & 0.223$\pm$0.002$\hspace*{1.6mm}$ & \textcolor{blue}{0.017$\pm$0.001}$\hspace*{1.6mm}$ & 0.289$\pm$0.002$\hspace*{1.6mm}$ & \textcolor{blue}{0.269$\pm$0.002}$\hspace*{1.6mm}$ & \textcolor{blue}{0.143$\pm$0.002}$\hspace*{1.6mm}$ \\					
		& Ours  & \textcolor{blue}{0.220$\pm$0.003}$\hspace*{1.6mm}$ & \textcolor{red}{0.016$\pm$0.001}$\hspace*{1.6mm}$ & \textcolor{blue}{0.291$\pm$0.002}$\hspace*{1.6mm}$ & \textcolor{red}{0.276$\pm$0.004}$\hspace*{1.6mm}$ & \textcolor{red}{0.149$\pm$0.001}$\hspace*{1.6mm}$ \\
		\hline
		\multirow{11}[10]{*}{tmc2007} & BR \cite{tsoumakas2007multi}    & 0.037$\pm$0.007$\bullet$ & 0.031$\pm$0.004$\bullet$ & 0.899$\pm$0.025$\bullet$ & 0.834$\pm$0.014$\bullet$ & 0.719$\pm$0.011$\bullet$ \\
		& LP \cite{boutell2004learning}    & 0.324$\pm$0.018$\bullet$ & 0.041$\pm$0.006$\bullet$ & 0.594$\pm$0.012$\bullet$ & 0.791$\pm$0.008$\bullet$ & 0.721$\pm$0.004$\bullet$ \\
		& ML-kNN \cite{zhang2007ml} & 0.031$\pm$0.006$\bullet$ & 0.058$\pm$0.004$\bullet$ & 0.844$\pm$0.017$\bullet$ & 0.682$\pm$0.003$\bullet$ & 0.493$\pm$0.002$\bullet$ \\
		& EPS \cite{read2008multi}   & 0.021$\pm$0.004$\bullet$ & 0.033$\pm$0.005$\bullet$ & 0.927$\pm$0.007$\bullet$ & 0.829$\pm$0.009$\bullet$ & 0.722$\pm$0.010$\bullet$ \\
		& ECC \cite{read2011classifier}  & 0.017$\pm$0.006$\bullet$ & 0.026$\pm$0.003$\bullet$ & 0.925$\pm$0.006$\bullet$ & 0.862$\pm$0.014$\bullet$ & 0.763$\pm$0.007$\bullet$ \\
		& RAkEL \cite{tsoumakas2011random} & 0.038$\pm$0.008$\bullet$ & 0.024$\pm$0.002$\bullet$ & 0.923$\pm$0.005$\bullet$ & 0.870$\pm$0.011$\bullet$ & 0.756$\pm$0.006$\bullet$ \\
		& CLR \cite{furnkranz2008multilabel}  & 0.018$\pm$0.005$\bullet$ & 0.034$\pm$0.004$\bullet$ & 0.923$\pm$0.011$\bullet$ & 0.825$\pm$0.013$\bullet$ & 0.711$\pm$0.011$\bullet$ \\
		& MLML \cite{hou2016multi} & 0.018$\pm$0.001$\bullet$ & 0.021$\pm$0.001$\bullet$ &  0.921$\pm$0.002$\bullet$ & 0.865$\pm$0.011$\bullet$ & 0.769$\pm$0.008$\bullet$ \\
		& MLFE \cite{zhang2018feature}    & 0.021$\pm$0.002$\bullet$ & 0.022$\pm$0.001$\bullet$ & 0.924$\pm$0.013$\bullet$ & 0.873$\pm$0.015$\bullet$ & 0.771$\pm$0.011$\bullet$ \\
		& HNOML \cite{zhang2019hybrid}   & 0.023$\pm$0.002$\bullet$ & 0.017$\pm$0.001$\bullet$ & 0.919$\pm$0.003$\bullet$ & 0.858$\pm$0.014$\bullet$ & 0.762$\pm$0.016$\bullet$ \\
		& Ours (linear)  & \textcolor{blue}{0.015$\pm$0.003}$\hspace*{1.6mm}$ & \textcolor{blue}{0.013$\pm$0.002}$\hspace*{1.6mm}$ & \textcolor{blue}{0.937$\pm$0.007}$\hspace*{1.6mm}$ & \textcolor{blue}{0.912$\pm$0.008}$\hspace*{1.6mm}$ & \textcolor{blue}{0.781$\pm$0.005}$\hspace*{1.6mm}$ \\					
		& Ours  & \textcolor{red}{0.012$\pm$0.004}$\hspace*{1.6mm}$ & \textcolor{red}{0.011$\pm$0.001}$\hspace*{1.6mm}$ & \textcolor{red}{0.945$\pm$0.007}$\hspace*{1.6mm}$ & \textcolor{red}{0.944$\pm$0.007}$\hspace*{1.6mm}$ & \textcolor{red}{0.792$\pm$0.010}$\hspace*{1.6mm}$ \\
		\bottomrule[1pt]
	\end{tabular}}%
	\label{tab:addlabel2}%
\end{table*}%

We compare our algorithm with both baseline and state-of-the-art multi-label classification methods. The binary relevance (BR) \cite{tsoumakas2007multi} and label powerset (LP) \cite{boutell2004learning} act as baselines. We also compare ours with two ensemble methods, \emph{i.e.}, ensemble of pruned sets (EPS) \cite{read2008multi} and ensemble of classifier chains (ECC) \cite{read2011classifier}, second-order approach - calibrated label ranking (CLR) \cite{furnkranz2008multilabel} and high-order approach - random k-labelsets (RAkEL) \cite{tsoumakas2011random}, the lazy multi-label methods based on k-nearest neighbors (ML-kNN)\cite{zhang2007ml} and feature-aware approach - multi-label manifold learning (MLML) \cite{hou2016multi}, labeling information enrichment approach - Multi-label Learning with Feature-induced labeling information Enrichment(MLFE) \cite{zhang2018feature} and robust approach for data with hybrid noise - hybrid noise-oriented multilabel learning (HNOML) \cite{zhang2019hybrid}.
We try our best to tune the parameters of all the above compared methods to the best performance according to the suggested ways in their literatures. 
\vspace{-0.4cm}
\begin{table*}[ht]
	\centering
	\renewcommand\arraystretch{0.7}
	\caption{Performance comparisons with approaches based on label space reduction.}
	\setlength{\tabcolsep}{1.0mm}{
		\begin{tabular}{cccccccc}
			\hline
			\multicolumn{2}{c}{Datasets} & \multicolumn{2}{c}{tmc2007} & \multicolumn{2}{c}{espgame} \\
			\hline
			Methods $\hspace*{2mm}/$ & Metrics & Micro-F1$\uparrow$ & Macro-F1$\uparrow$ & Micro-F1$\uparrow$ & Macro-F1$\uparrow$ \\
			\hline
			\multicolumn{2}{c}{MOPLMS \cite{Balasubramanian2012The}} & 0.556$\pm$0.012 & 0.421$\pm$0.013 & 0.032$\pm$0.006 & 0.025$\pm$0.005 \\
			\multicolumn{2}{c}{ML-CSSP \cite{bi2013efficient}} & 0.604$\pm$0.014 & 0.432$\pm$0.015 & 0.035$\pm$0.004 & 0.023$\pm$0.006 \\
			\multicolumn{2}{c}{PBR \cite{chen2012feature}} & 0.602$\pm$0.034 & 0.422$\pm$0.025 & 0.021$\pm$0.008 & 0.014$\pm$0.003 \\
			\multicolumn{2}{c}{CPLST \cite{chen2012feature}} & 0.643$\pm$0.027 & 0.437$\pm$0.031  & 0.042$\pm$0.005 & 0.023$\pm$0.004 \\
			\multicolumn{2}{c}{FAIE \cite{lin2014multi}} & 0.605$\pm$0.011 & 0.458$\pm$0.015  & 0.072$\pm$0.008 & 0.026$\pm$0.003 \\
			\multicolumn{2}{c}{Deep CPLST} & 0.786$\pm$0.021 & 0.601$\pm$0.031  & 0.074$\pm$0.004 & 0.016$\pm$0.002 \\
			\multicolumn{2}{c}{Deep FAIE} & 0.604$\pm$0.016 & 0.435$\pm$0.029  & 0.121$\pm$0.011 & 0.024$\pm$0.003 \\
			\multicolumn{2}{c}{LEML \cite{yu2014large}} & 0.704$\pm$0.013 & 0.616$\pm$0.022  & 0.148$\pm$0.004 & 0.082$\pm$0.001 \\
			\multicolumn{2}{c}{SLEEC \cite{bhatia2015sparse}} & 0.607$\pm$0.031 & 0.586$\pm$0.011  & 0.226$\pm$0.016 & 0.108$\pm$0.009 \\
			\multicolumn{2}{c}{DC2AE \cite{yeh2017learning}} & 0.808$\pm$0.017 & 0.757$\pm$0.027  & 0.256$\pm$0.013 & 0.121$\pm$0.009 \\
			\multicolumn{2}{c}{\textbf{Ours}} & \textbf{0.944$\pm$0.007} & \textbf{0.792$\pm$0.010}  & \textbf{0.276$\pm$0.004} & \textbf{0.149$\pm$0.001} \\
			\hline
		\end{tabular}%
		\label{tab:addlabel3}}%
\end{table*}%

As shown in Table \ref{tab:addlabel2}, we report the quantitative experimental results of different methods on the benchmark datasets. Because above comparison methods are not based on neural networks, for fair comparisons, we also report the results of our model using the linear projections instead of neural networks for feature embedding. For each algorithm, the averaged performance with standard deviation are reported in terms of different metrics. As for each metric,  ``$\uparrow$" indicates the larger the better  while  ``$\downarrow$" indicates the smaller the better. The red number and blue number indicate the best and the second best performances, respectively. According to Table \ref{tab:addlabel2}, several observations are obtained as follows: 1) Compared with other multi-label classification methods, our algorithm achieves competitive performance on all the five benchmark datasets. For example, on emotions , scene and tmc2007, our SPL-MLL ranks as the first in terms of all metrics. 2) Compared with BR and LP, our SPL-MLL obtains much better performance on all datasets. The reason may be that these methods lack of sufficient ability to explore complex correlations among labels. 3) Compared with the three ensemble methods EPS, ECC and RAkEL, our algorithm always performs better, which further verifies the effectiveness of our SPL-MLL. 4) We also note that the performances of ML-kNN, CLR and MLML are also competitive, and the performances of CLR are slightly better than ours on espgame in terms of some metrics. However, the performances of ours are more stable and robust for different datasets. For example, CLR performs unpromising on emotions, yeast, and scene.
5) Furthermore, compared with the latest and most advanced approaches MLFE and HNOML, our model outperforms them on all datasets in terms of most metrics. In short, our proposed SPL-MLL achieves promising and stable performance compared with state-of-the-art multi-label classification methods.

\begin{figure}[htbp]
	\centering
	\includegraphics[width=3.1in, height = 1.8in]{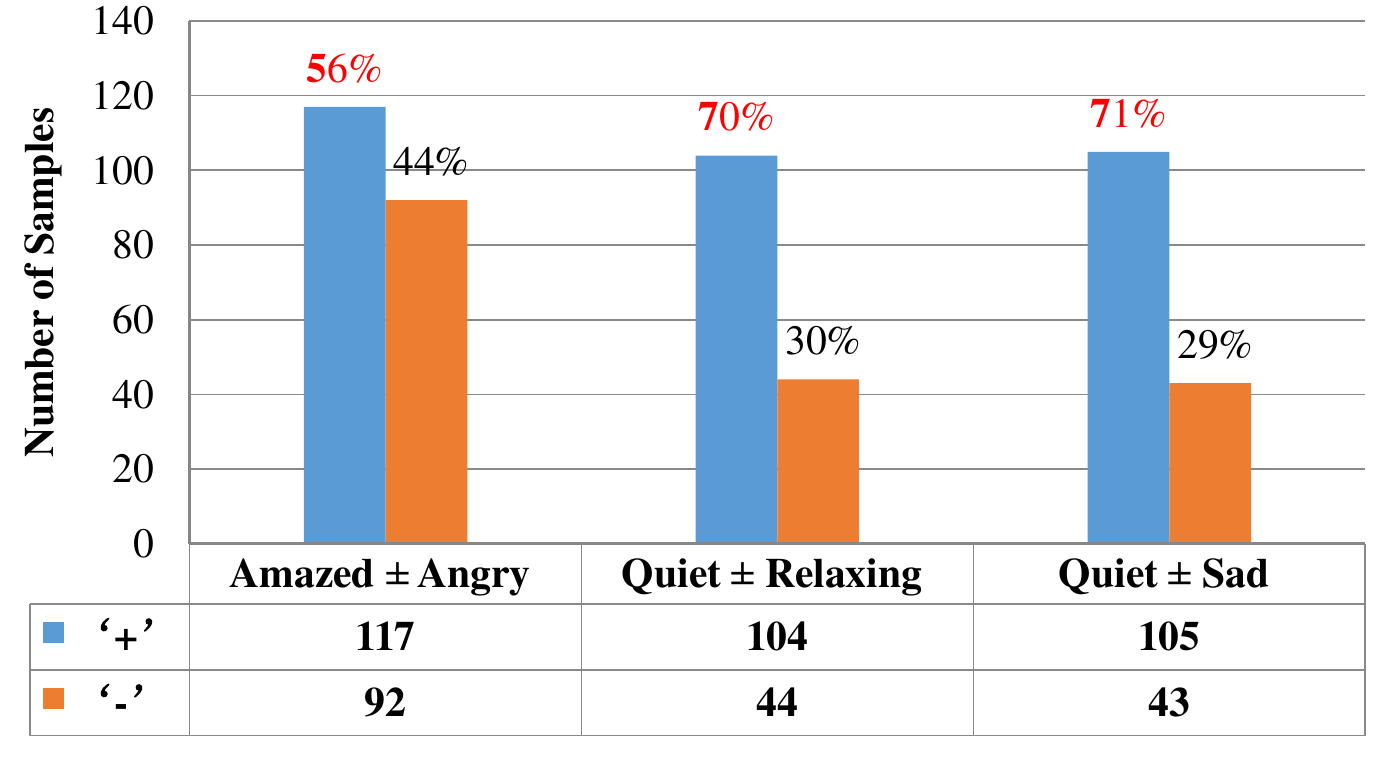}
	\caption{ Visualization of the number of co-occurrence labels on emotions. `+' and `-' denote co-occurrence and no co-occurrence of two labels, respectively.}
	\label{fig:Sample-emotions4}
\end{figure}
\vspace{-1.2cm}

\subsubsection{Comparison with label space reduction methods}
\begin{figure*}[!ht]
	\centering
	\subfigure[emotions]{
		\includegraphics[width=2.2in, height = 1.8in]{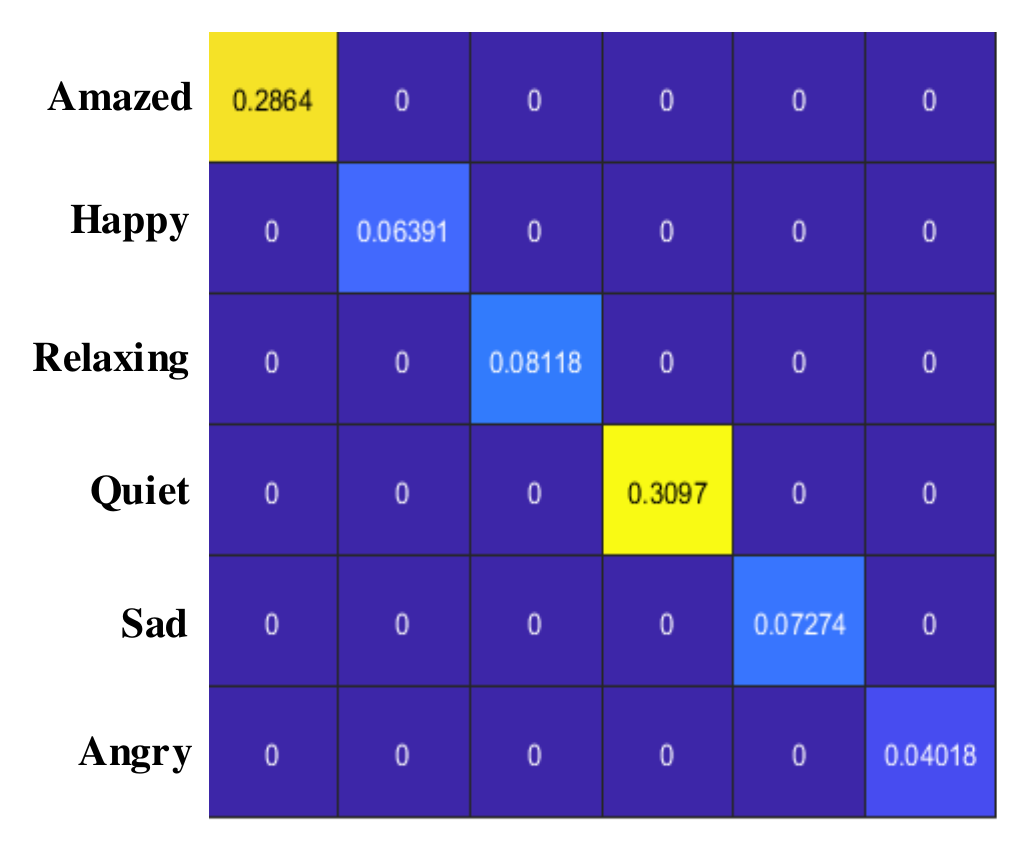}
	}
	\quad
	\subfigure[scene]{
		\includegraphics[width=2.2in, height = 1.8in]{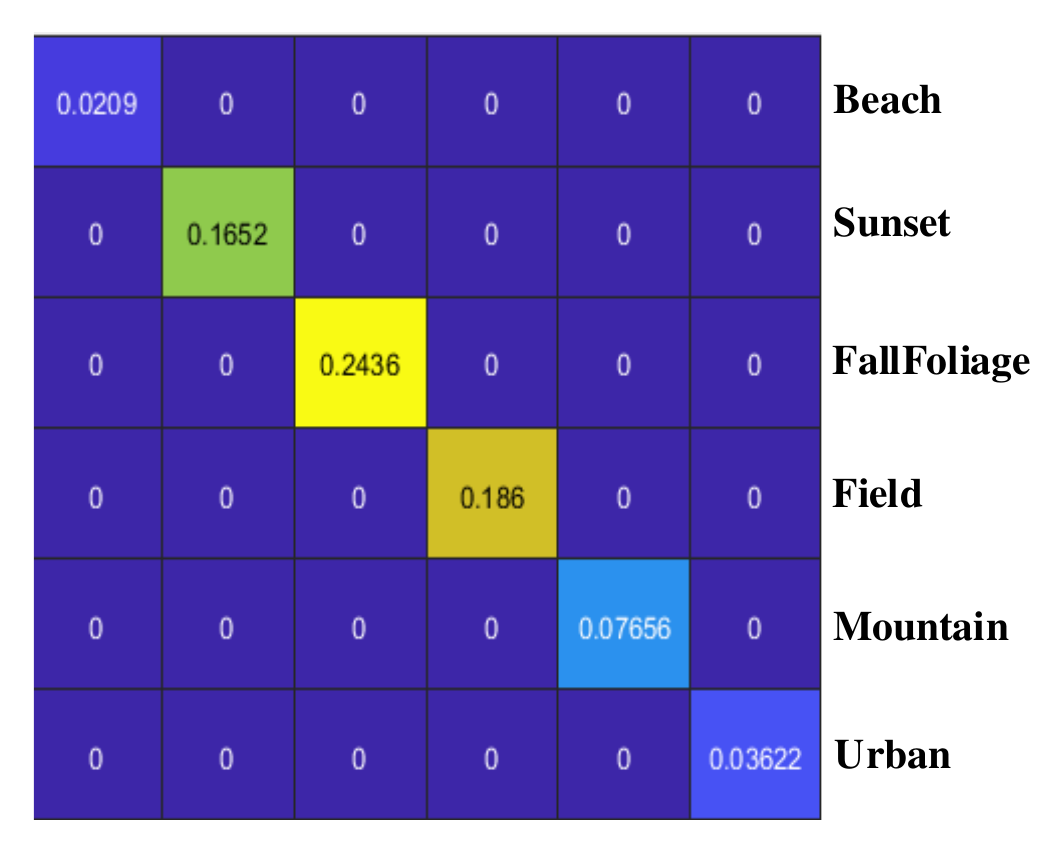}
	}
	\caption{Visualization of the landmark selection matrix $\mathbf{B}$.}
	\label{fig:B3}
\end{figure*}

We compare our method with two typical landmark selection methods which conduct landmark selection with group-sparsity technique (MOPLMS) \cite{Balasubramanian2012The} and an efficient randomized sampling procedure (ML-CSSP) \cite{bi2013efficient}. Moreover, we compare our method with the label embedding multi-label classification methods, which jointly reduce the label space and explore the correlations among labels. Specifically, we conduct comparison with the following label embedding based methods: Conditional Principal Label Space Transformation (CPLST) \cite{chen2012feature}, Feature-aware Implicit Label space Encoding (FaIE) \cite{lin2014multi}, Low rank Empirical risk minimization for Multi-Label Learning (LEML) \cite{yu2014large}, Sparse Local Embeddings for Extreme Multi-label Classification (SLEEC) \cite{bhatia2015sparse}, and the baseline method of partial binary relevance (PBR) \cite{chen2012feature}. Furthermore, we replace the linear regressors in CPLST and FAIE with DNN regressors, and name them as Deep CPLST and Deep FAIE, respectively.  The work in \cite{yeh2017learning} proposes a novel DNN architecture of Canonical-Correlated Autoencoder (C2AE), which can exploit label correlation effectively. Since some methods (e.g., C2AE) reported the results in terms of Micro-F1 and Macro-F1 \cite{tang2009large}, we also provides results of different approaches in terms of these two metrics for convenient comparison as shown in Table \ref{tab:addlabel3}. According to the results, it is observed that the performance of our model is much better than the landmark selection methods \cite{Balasubramanian2012The,bi2013efficient} which separate landmark selection and prediction in 2-step manner. Moreover, our SPL-MLL performs superiorly against these label embedding methods.
\vspace{-0.2cm}
\begin{table*}[!ht]\large
	\centering
	\renewcommand\arraystretch{1.1}
	\caption{Ablation studies for our model on different setting on pascal VOC 2007.}
	\resizebox{4.8in}{!}{
		\begin{tabular}{c|ccccc}
			\hline
			Methods & Ranking Loss $\downarrow$ & Hamming Loss $\downarrow$ & Average Precision $\uparrow$ & Micro-F1 $\uparrow$ & Macro-F1 $\uparrow$ \\
			\hline
			MLFE \cite{zhang2018feature}    & 0.232$\pm$0.013$\hspace*{1.6mm}$ & 0.162$\pm$0.012$\hspace*{1.6mm}$ & 0.565$\pm$0.022$\hspace*{1.6mm}$ & 0.436$\pm$0.026$\hspace*{1.6mm}$ & 0.357$\pm$0.011$\hspace*{1.6mm}$ \\
			HNOML \cite{zhang2019hybrid}   & 0.227$\pm$0.012$\hspace*{1.6mm}$ & 0.123$\pm$0.008$\hspace*{1.6mm}$ & 0.593$\pm$0.023$\hspace*{1.6mm}$ & 0.443$\pm$0.024$\hspace*{1.6mm}$ & 0.368$\pm$0.019$\hspace*{1.6mm}$ \\
			\hline
			NN-embeddings  & 0.324$\pm$0.016$\hspace*{1.6mm}$ & 0.266$\pm$0.011$\hspace*{1.6mm}$ & 0.431$\pm$0.013$\hspace*{1.6mm}$ & 0.308$\pm$0.011$\hspace*{1.6mm}$ & 0.287$\pm$0.016$\hspace*{1.6mm}$\\						
			Ours(NN + separated)  & 0.243$\pm$0.014$\hspace*{1.6mm}$ & 0.194$\pm$0.015$\hspace*{1.6mm}$ & 0.521$\pm$0.024$\hspace*{1.6mm}$ & 0.384$\pm$0.009$\hspace*{1.6mm}$ & 0.311$\pm$0.017$\hspace*{1.6mm}$ \\
			Ours(joint + linear) & 0.192$\pm$0.011$\hspace*{1.6mm}$ & 0.095$\pm$0.012$\hspace*{1.6mm}$ & 0.608$\pm$0.021$\hspace*{1.6mm}$ & 0.516$\pm$0.025$\hspace*{1.6mm}$ & 0.422$\pm$0.024$\hspace*{1.6mm}$ \\
			Ours(joint + NN)  & \textbf{0.184$\pm$0.012}$\hspace*{1.6mm}$ & \textbf{0.083$\pm$0.013}$\hspace*{1.6mm}$ & \textbf{0.616$\pm$0.018}$\hspace*{1.6mm}$ & \textbf{0.586$\pm$0.018}$\hspace*{1.6mm}$ & \textbf{0.495$\pm$0.017}$\hspace*{1.6mm}$ \\						
			\hline
	\end{tabular}}%
	\label{tab:addlabel4}%
\end{table*}%
\vspace{-0.4cm}

\subsubsection{Ablation Studies}
To investigate the advantage of our model on jointly conducting landmark selection, landmark prediction and label recovery in a unified 
framework, we further conduct comparison and ablation experiments on pascal VOC 2007. Specifically, we conduct ablation studies for our model under the following settings: (1) NN-embeddings: the features are directly encoded by neural network for full label recovery without landmark selection and landmark prediction; (2) Ours (NN + separated): our model is still based on the landmark selection strategy, but separates the landmark selection and landmark prediction in the 2-step manner like the work \cite{Balasubramanian2012The}; (3) Ours (joint + linear): our model employs the linear projections instead of neural networks for feature embedding. To further validate the performance improvement from our model, we also report the results of the latest and most advanced approaches MLFE \cite{zhang2018feature} and HNOML \cite{zhang2019hybrid}. The comparision results are shown in Table \ref{tab:addlabel4}, which validates the superiority of conducting landmark selection, landmark prediction and label recovery in a unified framework.

\subsubsection{Insight for selected landmarks}

To investigate the improvement of SPL-MLL, we visualize the landmark selection matrix $\mathbf{B}$ on emotions and scene. As illustrated in Fig.~\ref{fig:B3}, the values in yellow on the diagonal are much larger than the values in other colors, where the corresponding labels are selected landmarks. For emotions, ``Amazed" and ``Quiet" are most likely to be landmark labels, and ``Amazed" is often accompanied by ``Angry" in music, ``Quiet" tends to occur simultaneously with ``Relaxing" or ``Sad". Thus, we can utilize the selected landmark labels to recover other related labels effectively. Similar, for scene, the label ``FallFoliage" and ``Field" are most likely to be landmark labels.

As shown in Fig.~\ref{fig:Sample-emotions4}, we count the number of those samples with or without ``Angry" when having ``Amazed", which is represented as ``Amazed $\pm$ Angry", and similarly we obtain ``Quiet $\pm$ Relaxing" and ``Quiet $\pm$ Sad". According to Fig.~\ref{fig:Sample-emotions4}, it is observed that when the ``Amazed" (``Quiet") emotion occurs, the probability that ``Angry"  (``Relaxing" and ``Sad") occur simultaneously is $56\%$  ($70\%$ and $71\%$). This statistics further support the reasonability of the selected landmark labels.


\begin{figure*}[htbp]
	\centering
	\includegraphics[width=4.8in]{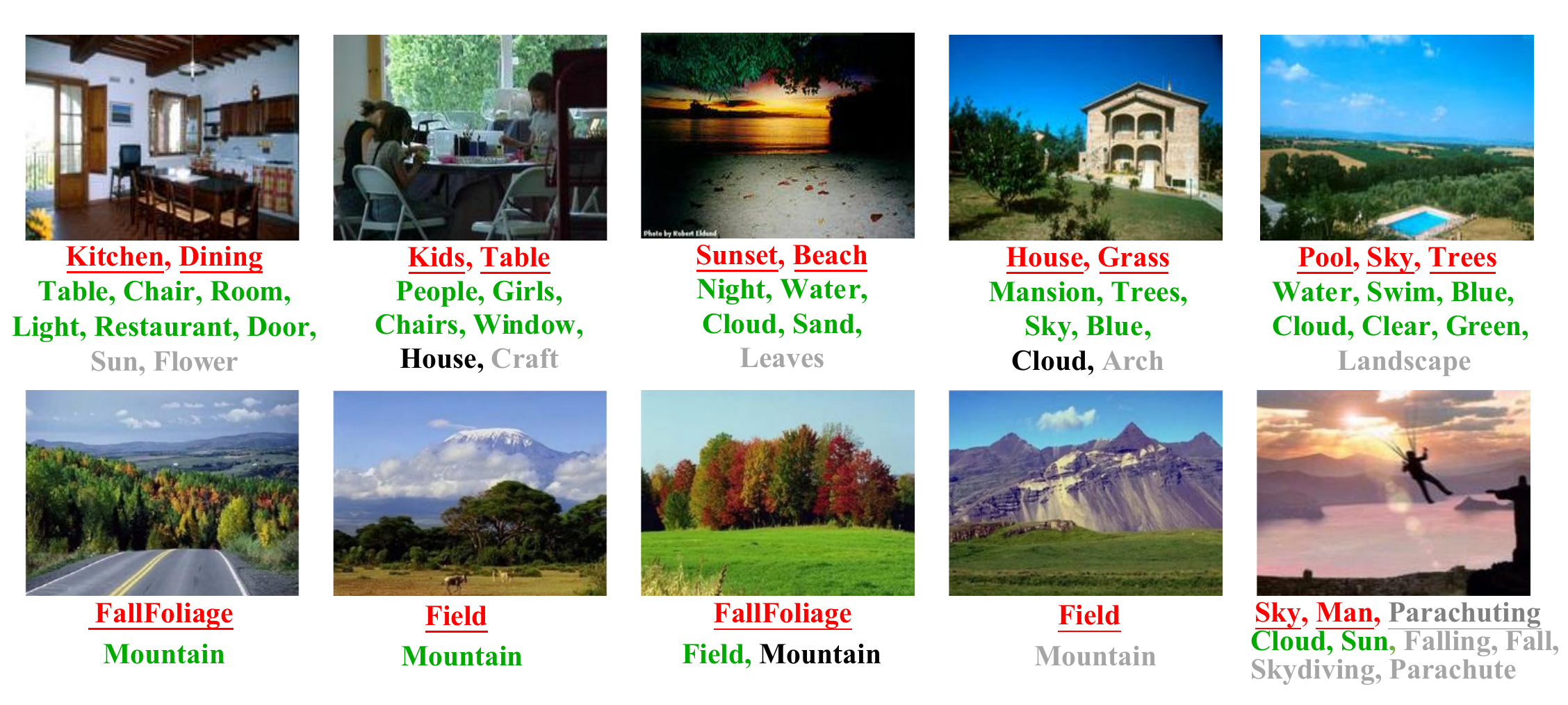}
	\caption{Example predictions  on espgame and scene. }
	\label{fig:images5}
\end{figure*}
\vspace{-0.3cm}
\subsubsection{Result visualization \& convergence experiment}
For intuitive analysis, Fig.~\ref{fig:images5} shows some representative examples from espgame and scene. The correctly predicted landmark labels from our model are in red, while the labels in green, gray and black indicate the successfully predicted, missed predicted and wrongly predicted labels. Generally, although multi-label classification is rather challenging especially for the large label set, our model achieves competitive results. We find that a few labels of some samples are not correctly predicted, and the possible reasons are as follows. First, a few labels on some samples do not obviously correlate with other labels, which makes it difficult to accurately recover given selected landmark labels. Second, a few labels for some samples are associated with very small parts in images, making it difficult to predict accurately even taking the feature of images into account in our model. For example, the image labeled with ``Kitchen" and ``Dining" as landmark labels has the following labels predicted correctly: ``Table", ``Chair", ``Room", ``Light", ``Restaurant" and ``Door". However, there are labels: ``Sun" and ``Flower" failed to be predicted. The main reasons is that the label ``Sun" and ``Flower" may be not strongly correlated with the selected landmark labels in the dataset.

There are a few landmarks failed to be predicted for some samples, even though our model aims to select predictable landmark labels. For example, for the rightmost picture in the bottom of Fig.~\ref{fig:images5}, we predict successfully ``Sky" and ``Man" as landmarks while not able to obtain the more critical landmark label ``Parachuting", which leads to failure prediction for ``Falling", ``Fall", ``Skydiving", ``Parachuting". It can be seen that the parachute is rather difficult to predict due to the strong illumination.

\begin{figure*}[!ht]
	\centering
	\subfigure[emotions]{
		\includegraphics[width=1.4in, height = 1.1in]{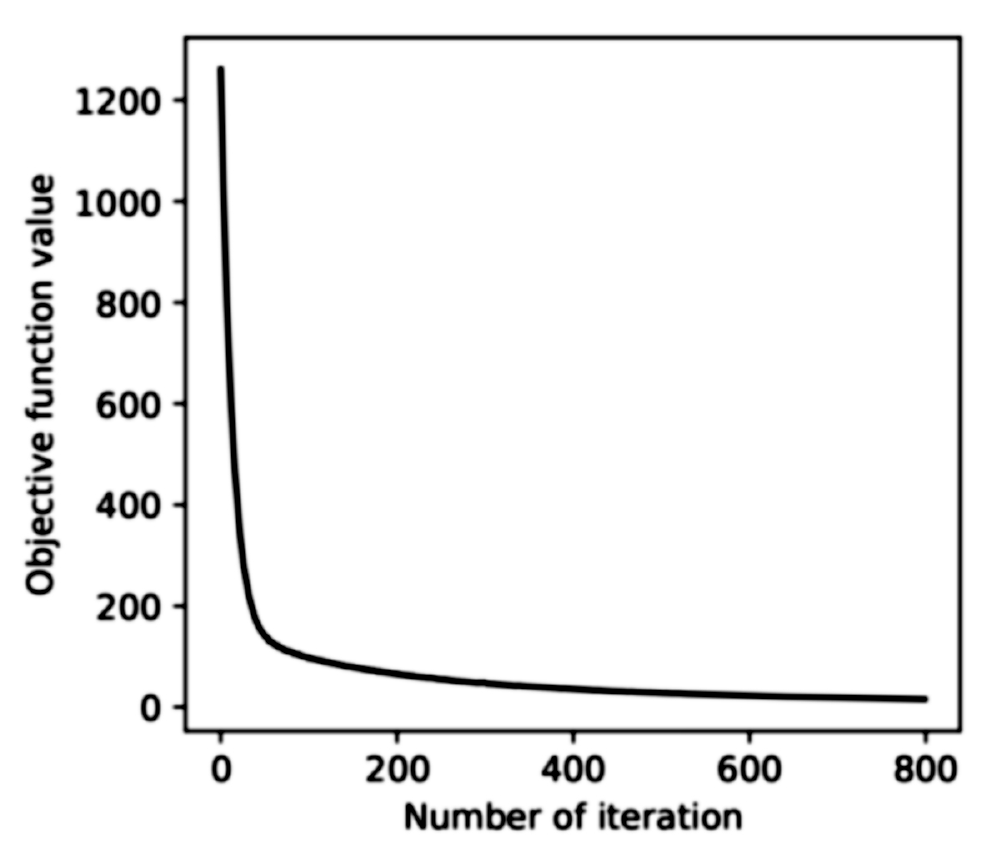}
	}
	\quad
	\subfigure[tmc2007]{
		\includegraphics[width=1.4in, height = 1.1in]{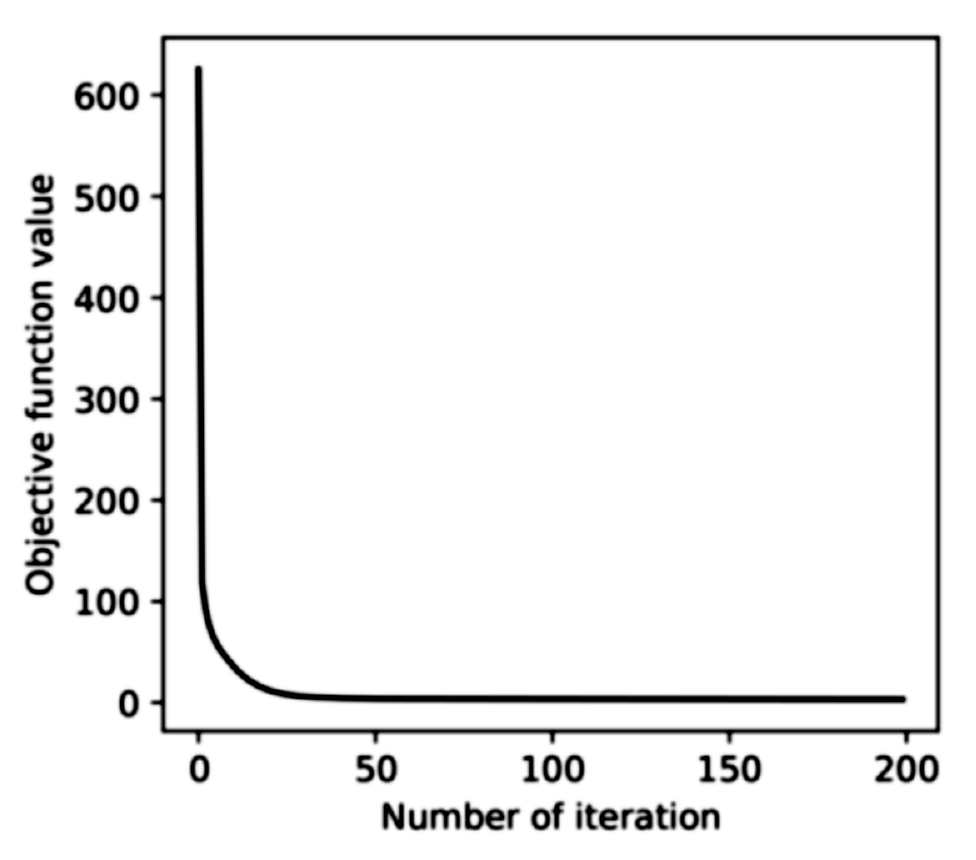}
	}
	\quad
	\subfigure[yeast]{
		\includegraphics[width=1.4in, height = 1.1in]{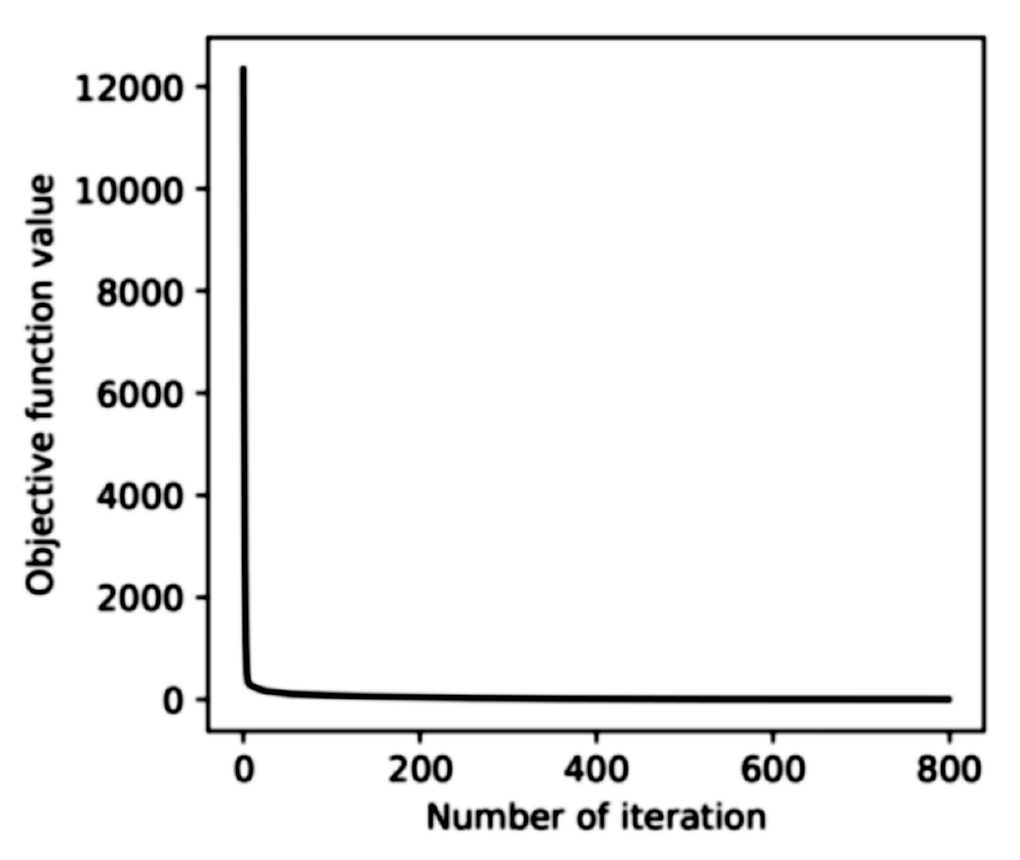}
	}
	\quad
	\caption{Convergence experiment.}
	\label{fig:loss-curve6}
\end{figure*}

Fig.~\ref{fig:loss-curve6} gives the convergence experiments on emotion, yeast and tmc2007. Obviously, the results demonstrate that our method can converge within a small number of iterations.

\section{Conclusions \& Future Work}
In this paper, we proposed a novel landmark-based multi-label classification algorithm, termed \emph{SPL-MLL: Selecting Predictable Landmarks for Multi-Label Learning}. SPL-MLL jointly takes the representative and predictable properties for landmarks in a unified framework, avoiding separating landmark selection/prediction in the 2-step manner. Our key idea lies in selecting explicitly the landmarks which are both representative and predictable. The empirical experiments clearly demonstrate that our algorithm outperforms existing state-of-the-art methods. In the future, we will consider the end-to-end manner to extend our model for image annotation with large label set.

\subsubsection{Acknowledgements.} This work is supported by the National Natural Science Foundation of China (Nos. 61976151, 61732011 and 61872190).

%
%
\bibliographystyle{splncs04}
\bibliography{egbib}
\end{document}